\documentclass[letterpaper, conference]{ieeeconf}
\IEEEoverridecommandlockouts                                      %
\usepackage[utf8]{inputenc}
\usepackage{amsmath,amssymb,amsfonts}
\usepackage{bm}
\usepackage{url}
\usepackage{graphicx,graphics}
\usepackage{booktabs}
\usepackage{multirow}
\usepackage[ruled,linesnumbered]{algorithm2e}
\usepackage{textcomp}
\usepackage{tabularx}
\usepackage[flushleft]{threeparttable}
\usepackage{cite}
\usepackage[table,dvipsnames]{xcolor}
\usepackage{soulutf8}
\usepackage{soul}
\usepackage{amssymb}
\usepackage{algpseudocode}

\newcommand{\midsepremove}{\aboverulesep = 0.2mm \belowrulesep = 0.2mm}
    \midsepremove
\newcommand{\midsepdefault}{\aboverulesep = 0.605mm \belowrulesep = 0.984mm}
    \midsepdefault

\title{\LARGE \bf 3D Spectral Domain Registration-Based Visual Servoing}
\author{{Maxime Adjigble$^{\dagger*}$, Brahim Tamadazte$^{\ddagger}$, Cristiana de Farias$^{\dagger}$, Rustam Stolkin$^{\dagger}$, Naresh Marturi$^{\dagger}$}%
\thanks{$^{\dagger}$Extreme Robotics Laboratory, School of Metallurgy and Materials, University of Birmingham, Edgbaston, B15 2TT, UK. $^{\ddagger}$Sorbonne Universit\'e, CNRS UMR 7222, INSERM U1150, ISIR, F-75005, Paris, France.}%
    \thanks{This research was funded by the EPSRC under grant EP/P01366X/1, and in part supported by CHIST-ERA under Project EP/S032428/1 PeGRoGAM and by the Faraday Institution [grant number FIRG005].}%
    \thanks{$^{*}$Corresponding Author: \texttt{m.k.j.adjigble@bham.ac.uk}}%
}

\begin{document}\sloppy
\bstctlcite{IEEEexample:BSTcontrol}
\maketitle

\begin{abstract}
This paper presents a spectral domain registration-based visual servoing scheme that works on 3D point clouds. Specifically, we propose a 3D model/point cloud alignment method, which works by finding a global transformation between reference and target point clouds using spectral analysis. A 3D Fast Fourier Transform (FFT) in $\mathbb{R}^3$ is used for the translation estimation, and the real spherical harmonics in $\bm{SO(3)}$ are used for the rotations estimation. Such an approach allows us to derive a decoupled 6 degrees of freedom (DoF) controller, where we use gradient ascent optimisation to minimise translation and rotational costs. We then show how this methodology can be used to regulate a robot arm to perform a positioning task. In contrast to the existing state-of-the-art depth-based visual servoing methods that either require dense depth maps or dense point clouds, our method works well with partial point clouds and can effectively handle larger transformations between the reference and the target positions. Furthermore, the use of spectral data (instead of spatial data) for transformation estimation makes our method robust to sensor-induced noise and partial occlusions. We validate our approach by performing experiments using point clouds acquired by a robot-mounted depth camera. Obtained results demonstrate the effectiveness of our visual servoing approach.
\end{abstract}

\section{Introduction}
\label{sec:intro}
The last three decades have seen a growing focus on visual servoing methods to perform robotic tasks in various sectors, e.g., industry, defence, autonomous vehicles, aerospace, medicine etc. \emph{Visual servoing} refers to the dynamic control of systems using continuous visual feedback. Consequently, the key components of a classical visual servoing controller are feature extraction, matching, and tracking over time using images. Nevertheless, the feasibility and effectiveness of classical visual servoing are closely correlated with that of visual tracking methods, whose performance is a concern with low-textured images that do not have distinguishable geometric shapes.

Recently, advanced visual servoing methods emerged that allow avoiding visual tracking by directly using global image information for error regulation in the control loop. These approaches are referred to as \emph{direct visual servoing} methods \cite{deguchi2000direct}. Accordingly, different types of global image information have been investigated in the literature such as image-intensity~\cite{Tamadazte2012,collewet2011photometric}, spatio-temporal gradients \cite{Marchand2010gradient}, histograms \cite{bateux2016histograms}, mutual information \cite{dame2011mutual}, Gaussian mixtures \cite{crombez2018visual} etc. More recently, authors proposed to model time-frequency image information such as wavelets~\cite{ourak2019direct} and shearlets~\cite{duflot2019wavelet} instead of the spatio-temporal image information. However, direct methods clearly exhibit narrower convergence domains compared to the traditional visual servoing schemes. To tackle this problem, some works have used spectral domain visual features. These features are proved to be robust to noise and are used for many computer vision and robotics applications like image correlation \cite{larsson2011correlating}, range data registration \cite{bulow2012spectral}, robotic grasping \cite{adjigble2021spectgrasp} etc. In~\cite{marchand2020direct}, Discrete Cosine Transform (DCT) coefficients have been explored for direct visual servoing and in~\cite{marturi2014visual, marturi2016image}, Fourier shift property has been used to design a decoupled visual controller.
\begin{figure}
    \centering
    \includegraphics[width=\columnwidth]{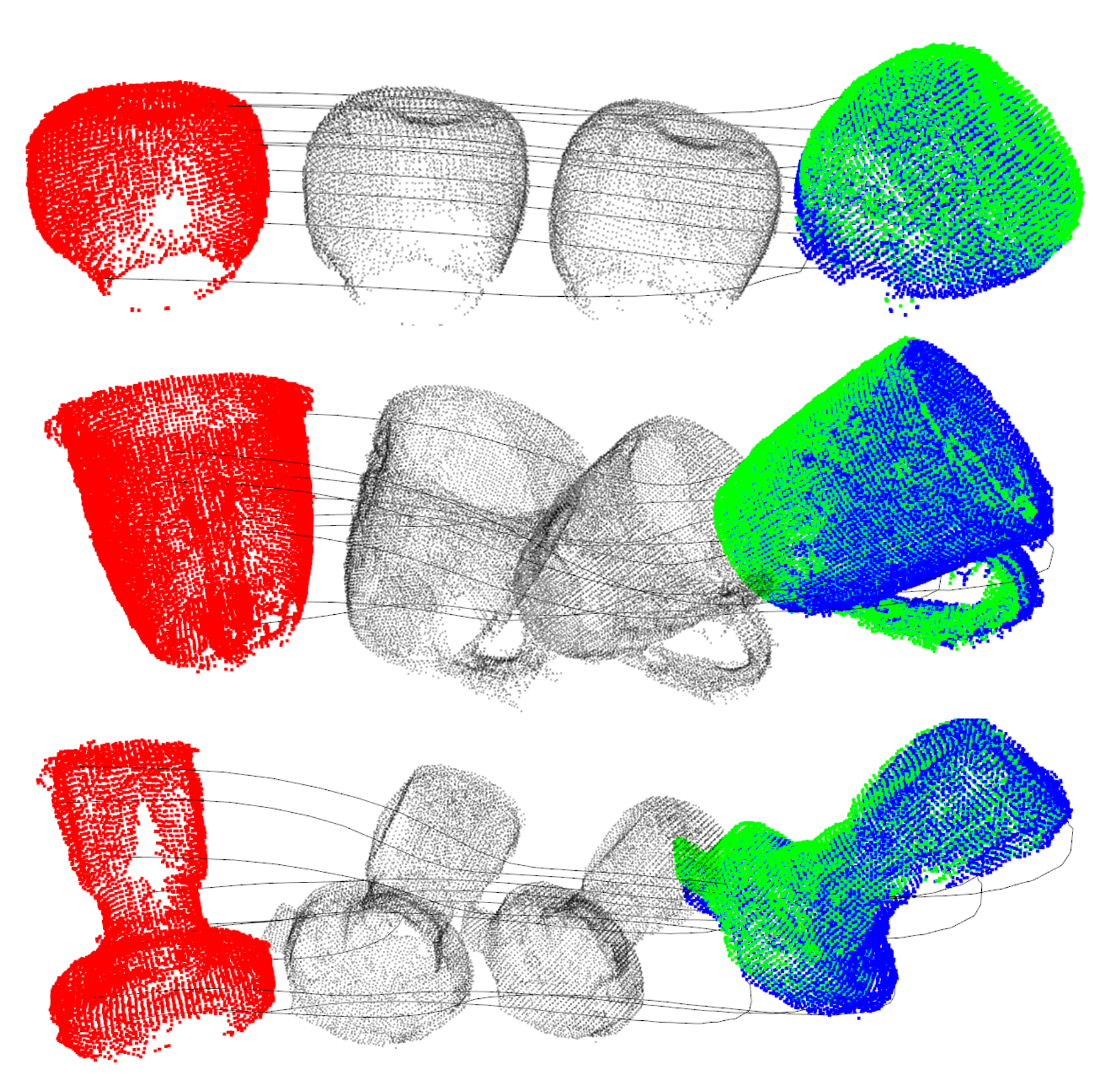}
    \caption{Illustration of the model alignment process with our proposed approach for 3 different objects: (top) apple; (middle) mug; and (bottom) gas knob. Red point cloud is the reference model, gray ones are intermediate candidates during the convergence, green is the target scene point cloud and blue is the final aligned model with the target. The gray trajectory curves indicate the model convergence.}
    \label{fig:fig1}
\end{figure}

Most of the aforementioned state of the art on visual servoing is based on 2D image information and the literature using 3D data for visual servoing, e.g., depth maps or point clouds, is very much limited. Very few recent works have reported such methods \cite{teuliere2014dense, 8122920, dahroug2020pca}. A particular advantage of using 3D data over 2D images is that they are well-suited for complex environments, \textit{i.e.}, texture-less, varying light, unstructured etc., and avoid computation of complex pose estimations. A direct visual servoing method based only on camera-acquired depth maps is presented in~\cite{teuliere2014dense}. The control law minimises the depth error computed using the current and reference full-depth maps. A similar approach has been reported in~\cite{8122920} to control the motion of a mobile robot. Although these methods reported promising results, they require dense depth data and exhibit limited convergence. A virtual visual servoing method using a polygon mesh generated offline from point clouds is presented in~\cite{kingkan2016model}. Although point clouds are used in its offline phase, the method's main visual controller still uses stereo image pairs for model matching.

In this paper, we present a spectral domain registration-based visual servoing scheme using point clouds. Although, very few works have used spectral information in their 2D visual servoing schemes \cite{marchand2020direct, ourak2016wavelets, marturi2016image, marturi2014visual}, there is no known instance reported in the literature that uses spectrally transformed point clouds. The main basis of our approach is the 3D model/point cloud alignment or registration, which works by finding a global 6 degrees of freedom (DoF) transformation between the two point clouds corresponding respectively to a reference model and a target object or scene. Using spectrally transformed point clouds, translation is estimated by Fourier analysis whereas rotation is estimated by spherical correlation. Gradient ascent-based optimization has then been used to iteratively minimize translation and rotation costs. Object translations and rotations are considered independent. This allows us to devise a decoupled control scheme, which optimizes the two costs separately. The proposed method uses a 3D fast Fourier transform in the Cartesian space $\mathbb{R}^3$ and real spherical harmonics on the unit-sphere $\bm{S}^2$ and the rotation group $\bm{SO(3)}$ to compute the gradient of the translation and rotation costs, respectively. This method can be used for aligning the entire global scene (direct visual servoing) or a single object model in a simple or cluttered scene. Example instances of object alignments in simple scenes are shown in Fig.~\ref{fig:fig1}, where a reference model is being aligned with a (single object) scene point cloud. While the current or the target scene cloud is captured online by a scene or robot-mounted depth sensor, the reference cloud can be a transformed global scene cloud (in case of direct visual servoing) or can be obtained offline by sampling a CAD model of the target object or by registering multiple cloud samples of the object as in~\cite{marturi2019dynamic}.

The key contributions of this work are as follows:
\begin{itemize}
    \item We propose a new spectral domain-based method for full 3D model alignment, \textit{i.e.}, to estimate global translation and rotation between two point clouds.
    \item We propose a new 6-DoF visual servoing scheme that works directly with dense as well as partial point clouds represented in the spectral domain.
\end{itemize}

Our method's advantages are multi-fold. Unlike the existing approaches that require dense data, our method can work effectively with partial point clouds. In comparison to the state-of-the-art depth-based direct visual servoing schemes, our method possesses improved convergence domain. The use of spectral data instead of spatial data makes our method robust to noise, which is apparent when using real-world point cloud measurements. Since no colour or intensity information is required, our method can work well in the case of texture-less objects and low lighting conditions. Finally, the proposed method can be used for both aligning an object model in a densely cluttered scene, and positioning a robot manipulator in the task space.%
%
\section{Methodology}
\label{sec:method}
In this section, we present our proposed 3D spectral domain visual servoing method. As mentioned earlier, the main basis of our approach is the model registration schema using spectrally transformed point clouds. To this extent, we first introduce the representation used by our method followed by the concepts of phase correlation in the Cartesian space and on the unit-sphere. Finally, the derived control law is presented.
\subsection{Point Cloud Representation}
The first step of our 3D visual servoing pipeline is to represent the points and the surface normals of the point cloud as a voxel grid and as an Extended Gaussian Image (EGI), respectively.
\subsubsection{Points as voxel grid}
The discretisation of a point cloud is a straightforward process. Given a point cloud composed of $N$ points, a 3D voxel grid of resolution $r \in \mathbb{R}^+$ can be constructed. For each point $p = (x, y, z)$ of the point cloud, the voxel indices of the point $p_{ijk}= (i, j, k)$ are computed as:
\begin{equation}
\label{eq:voxel_index}
i = [x/r]\qquad
    j = [y/r]\qquad
    k = [z/r]
\end{equation}
where, the operation $[./.]$ represents integer division, \textit{i.e.}, only the integer part of the division is retained.

Let $v_t: \mathbb{R}^3 \rightarrow \mathbb{N}^3$ be the mapping between the Cartesian coordinates and the voxel indices. The voxel grid function\footnote{The subscript $t$ indicates that the function is used for translation estimation, in the same way, the subscript $r$ will be used for functions related to rotation estimation.} $f_t : \mathbb{R}^3 \rightarrow \mathbb{N}$ of a point cloud can be defined as:
\begin{equation}
f_t(p) = f_t(x, y, z) = v_{ijk}
\label{eq:voxel}
\end{equation}
where, $v_{i,j,k} \in [0,1]$. Here, $v_{i,j,k}=1$ if at least one point of the point cloud has indices equal to $v_T(p) = (i,j,k)$ and $v_{ijk} = 0$ otherwise. Our method uses a voxel grid with binary values; however, a voxel grid with real values can also be used in the same way. The Local Contact Moments (LoCoMo) metric presented in~\cite{adjigble2018model} can be a good candidate for enhancing the information contained in the voxel grid.

\subsubsection{Surface normals as Extended Gaussian Image}
The EGI is a popular representation of functions expressed in the unit-sphere. It has extensively been used in the literature as a shape descriptor for object surface normals \cite{little1985extended,nayar1990specular,lowekamp2002exploring,adjigble2021spectgrasp}. Changing the representation of a surface normal $n = (n_x, n_y, n_z) \in \mathbb{R}^3$ from Euclidean to spherical coordinates $n = (r, \theta, \phi)$ using \eqref{eq:spherical_coordinates}, allows expressing the surface normal on the unit-sphere. 
\begin{equation}
\label{eq:spherical_coordinates}
\begin{gathered}
    r      = \sqrt{n_x^2 + n_y^2 + n_z^2} \qquad
    \theta = \arctan{\frac{\sqrt{n_x^2 + n_y^2}}{n_z}} \\
   \phi    = \arctan(\frac{n_y}{n_x}) 
\end{gathered}
\end{equation}

\begin{figure}
    \centering
    \includegraphics[width=\columnwidth]{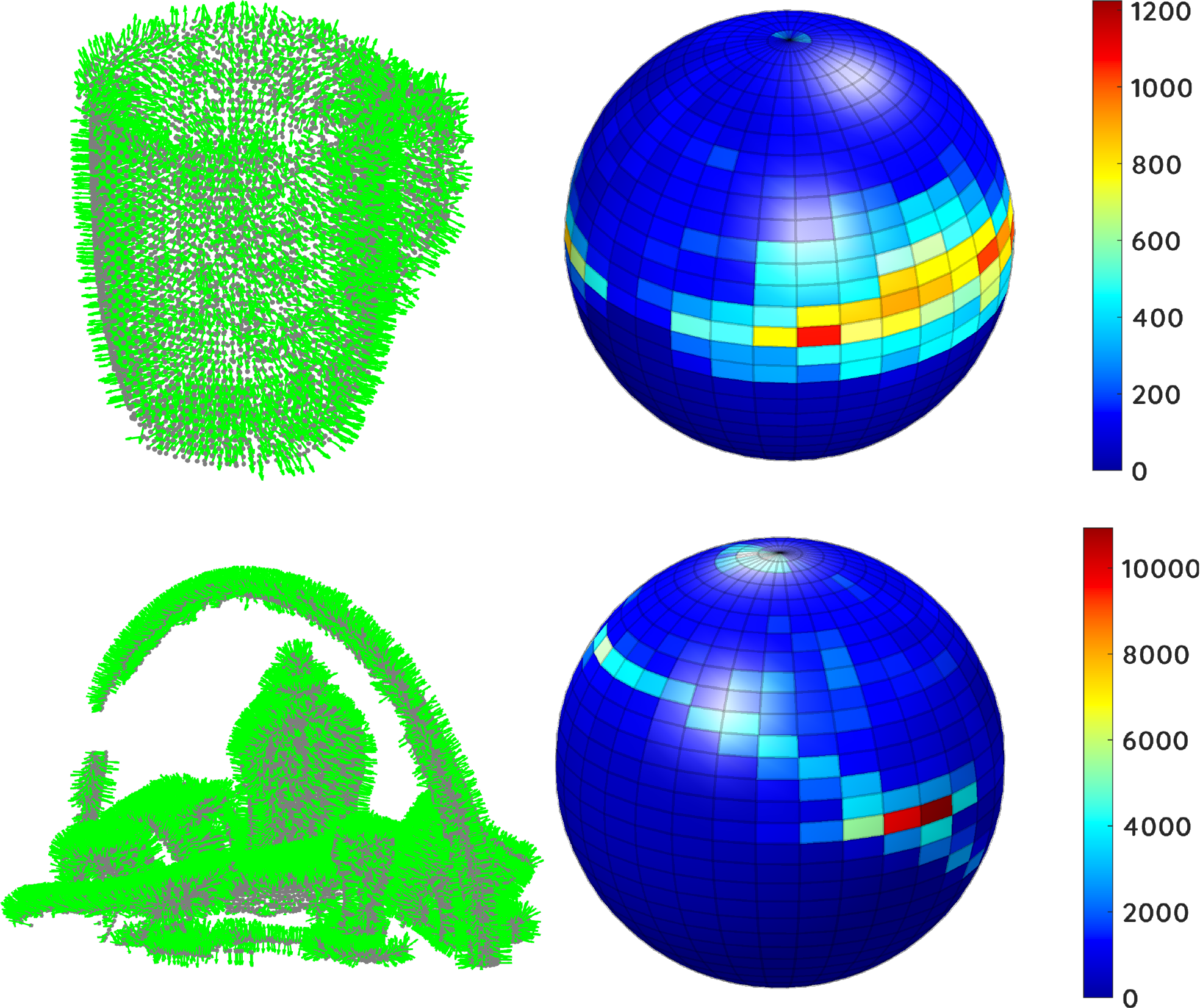}
    \caption{EGIs of (top) "mug" object and (bottom) cluttered scene, which are represented as point clouds with surface normals (small green arrows).}
    \label{fig:fig2_egi}
\end{figure}

The radial distance $r = 1$ for all surface normals as they are unitary vectors. Thus, the set $(\theta, \phi)$ is sufficient to describe the distribution of surface normals on the unit-sphere.
A discrete representation of the sphere is required to perform numerical computations. The following discretisation along the longitude and latitude is used: $\theta_j = \frac{\pi(2j + 1)}{4B}$ and $\phi_k = \frac{\pi k}{B}$  , $(j, k) \in \mathbb{N}$ with the constraint $0  \leq j, k < 2B$ and $B \in \mathbb{N}$ is the bandwidth. The value of the bandwidth is usually chosen as a power of $2$, meaning that $B=2^n, n \in \mathbb{N}^+$.
The EGI of the surface normals of a given point cloud can then be expressed as the function $f_r : \bm{S}^2 \rightarrow \mathbb{N}$:
\begin{equation}
f_r(\theta, \phi) = c_r({\theta_j, \phi_k})
\label{eq:egi}
\end{equation}
where, $c_r \in \mathbb{N}$ is the count that represents the number of surface normals in the point cloud with discrete longitude and latitude equal to $(\theta_j, \phi_k)$. In this case, count values are used instead of binary values. The advantage is that a distribution of surface normals on the unit-sphere provides more information on the geometry of the object compared to a simple binary distribution. Fig.~\ref{fig:fig2_egi} shows sample EGIs of an object and a clutter scene represented as point clouds.
%
\subsection{Translation Estimation via Fourier Analysis on $\mathbb{R}^3$}
The translation between the target and reference point clouds is estimated using 3D phase correlation in the spectral domain with Fourier analysis. The main advantage is that Fourier analysis-based methods are robust to noisy measurements \cite{bulow2012spectral,marturi2016image}. The phase correlation method is based on the Fourier shift property and maps translations in the Cartesian space to phase shift in the spectral domain. 

 Let $f_t : \mathbb{R}^3 \rightarrow \mathbb{N}$ be the voxel representation of the point cloud of an object or a scene. The Fourier coefficients of $f_t$ are computed as:
\begin{equation}
F_t(u, v, w) = \sum_{x=0}^{M-1}\sum_{y=0}^{N-1}\sum_{z=0}^{L-1}f_t(x,y,z)e^{-i2\pi(\frac{u}{M}x + \frac{v}{N}y + \frac{w}{L}z)}
\label{eq:fourier_coefs}
\end{equation}
where, $M, N, L \in N^+$ are the maximum degree of expansion of the Fourier coefficients in the $X$, $Y$, and $Z$ axes, respectively and $(u, v, w)$ are the frequency domain coordinates. Suppose the object or scene is translated by $T = (\tau_x,\tau_y, \tau_z) \in \mathbb{R}^3$, and let $g_t : \mathbb{R}^3 \rightarrow \mathbb{N}$ be the new voxel representation of the translated point cloud. Based on the Fourier shift property, the Fourier coefficients $G_t$ of $g_t$ can be computed by:
\begin{equation}
G_t(u, v, w) = F_t(u, v, w)e^{-i2\pi(\frac{u}{M}\tau_x + \frac{v}{N}\tau_y + \frac{w}{L}\tau_z)}
\label{eq:fourier_shift_theorem}
\end{equation}

The aim of the translation estimation is to find $T$ given $f_t$ and $g_t$. This can be efficiently done by first computing the normalized cross-power spectrum $\mathcal{C}_t$ of $F_t$ and $G_t$, and applying the inverse Fourier transform by \eqref{eq:cross_power_spectrum}. 
\begin{equation}
\begin{aligned}
\mathcal{C}_t(u,v,w) &= \frac{F_t(u,v,w)\overline{G_t(u,v,w)}}{|F_t(u,v,w)\overline{G_t(u,v,w)}|} \\
\delta(\tau_x,\tau_y, \tau_z) &= \mathcal{F}^{-1}(\mathcal{C}_t(u,v,w)) 
\label{eq:cross_power_spectrum}
\end{aligned}
\end{equation}
where, $\overline{G_t}$ is the complex conjugate of $G_t$, and $\mathcal{F}^{-1}$ is the inverse Fourier transform. The result $\delta$ is the \emph{Dirac delta function} whose peak location corresponds to the translation $T$. Therefore, $T$ can be found by maximizing the $\delta$.

\begin{equation}
\begin{aligned}
T &= \nabla_{glob} T = \mathrm{argmax}\{\delta(\tau_x,\tau_y, \tau_z)\}
\label{eq:fourier_shift_solution}
\end{aligned}
\end{equation}
Even though the global solution $\nabla_{glob} T$ for the translation can be found directly, in the context of 3D visual servoing, only a small step $\nabla T = \lambda_t\nabla_{glob} T$, with $\lambda_t \in \mathbb{R}^+$ and $\lambda_t < 1$, will be taken at each iteration. This allows the translation and rotation to be estimated concurrently, but also to control the dynamics of the controller. The following cost-function $J_t(T)$ can be formulated to evaluate the performance of the translation estimation algorithm on $\mathbb{R}^3$:
\begin{equation}
\begin{aligned}
J_t(T) = \frac{1}{2}||g_t(x) - f_t(x + T)||^2 \\
\end{aligned}
\label{eq:translation_cost_function}
\end{equation}
\subsection{Rotation Estimation via Fourier Analysis on $\bm{S}^2$}
Similarly to the translation, the rotation between the target and reference point clouds can also be estimated by using spectral analysis. Here, the unitary representation of signals expressed on the unit-sphere is used to encode the information of the object's surface normals. The same advantage as for the translation estimation applies, \emph{i.e.}, robustness to noise. In this case, we estimate the global rotation via EGI correlation. It is possible to find the optimal rotation directly by searching for the rotation maximizing the correlation, but this involves performing a double integration which can be computationally expensive. Instead, the analytical gradient of the correlation is used to iteratively compute the rotation that maximizes the correlation.
\subsubsection{Fourier transform on $\bm{S}^2$ and $\bm{SO}(3)$}
Let $f_r : \bm{S}^2 \rightarrow \mathbb{N}$ be the EGI of the surface normals of an object. Because $f_r$ has values in $\mathbb{N} \subset \mathbb{R}$, real harmonic analysis on $\bm{SO(3)}$, introduced in~\cite{lee2018real}, can be used to compute the Fourier parameters. Given a bandwidth $B$, the Fourier transform of $f_r$ on $\bm{S}^2$ is expressed as:
\begin{equation}
f_r(\theta, \phi) = \sum_{l=0}^{B-1}(F^l_r)^TS^l(\theta, \phi)
\label{eq:real_fourier_transform_fr}
\end{equation}
where, $F^l_r \in \mathbb{R}^{(2l+1)\times1}$ are the Fourier parameters and $S^l \in \mathbb{R}^{2l+1}$ are the orthogonal basis for real-value functions on $\bm{S}^2$. The vector $S^l$ is constructed from the real spherical harmonics $Y^l(\theta, \phi$), and a matrix $T^l \in \mathbb{C}^{(2l+1)\times(2l+1)}$ of complex coefficients as:
\begin{equation}
S^l(\theta, \phi) = T^lY^l(\theta, \phi)
\label{eq:real_fourier_transform}
\end{equation}
Refer \cite{lee2018real, blanco1997evaluation} for more details on spherical harmonics.

Let us suppose that the point cloud is rotated around its center of mass by a rotation $R \in SO(3)$ parameterized by the $ZYZ$ Euler angles $\alpha, \gamma \in [0,2\pi[$ and $\beta \in [0, \pi]$, with $g_r : \bm{S}^2 \rightarrow \mathbb{N}$ being the EGI of the rotated point cloud. Thereby, the rotation matrix $R$ can be expressed as:
\begin{equation}
R = R(\alpha, \beta, \gamma) = \exp(\alpha\hat{e}_z)\exp(\beta\hat{e}_y)\exp(\gamma\hat{e}_z)
\label{eq:euler_representation}
\end{equation}
where, $e_y$ and $e_z$ are the vectors $(0,1,0)$ and $(0,0,1)$, respectively. The operator $\hat{.}: \mathbb{R}^3 \rightarrow \mathfrak{so}(3)$ transforms a 3D vector into its $3\times3$ skew-symmetric matrix via the Lie algebra $\mathfrak{so}(3) = \{S \in R^{3\times3} | S + S^T = 0\}$.
Even though the representation in \eqref{eq:euler_representation} presents inherent singularities, it is extremely convenient for the computation of the Fourier transform on $\bm{SO(3)}$. Same as in \eqref{eq:real_fourier_transform_fr}, the Fourier transform of $g_r$ is given as:
\begin{equation}
g_r(\theta, \phi) = \sum_{l=0}^{B-1}(G^l_r)^TS^l(\theta, \phi) 
\label{eq:real_fourier_transform_gr}
\end{equation}
where $G^l_r \in \mathbb{R}^{(2l+1)\times1}$ are the Fourier parameters.

Considering that $g_r$ is a rotated version of $f_r$ and thus~\eqref{eq:gr} holds, the Fourier transform of $g_r$ can be computed using the Fourier parameter of $f_r$ by~\eqref{eq:real_fourier_transform_gr_simplified}.
\begin{equation}
g_r(\theta, \phi) = f_r(R^T(\theta, \phi))
\label{eq:gr}
\end{equation}
$R^T(\theta, \phi)$ is a notation shortcut for the expression $M_{s2c}^{-1}(R^TM_{s2c}(\theta, \phi))$, where $M_{s2c}:\bm{S}^2 \rightarrow \mathbb{R}^3$ is the function converting spherical to Cartesian coordinates, and $M_{s2c}^{-1}:\mathbb{R}^3 \rightarrow \bm{S}^2$, its inverse can be obtained using~\eqref{eq:spherical_coordinates}. Rewriting \eqref{eq:gr},
\begin{equation}
\begin{aligned}
g_r(\theta, \phi) &= \sum_{l=0}^{B-1}(F^l_r)^TS^l(R^T(\theta, \phi)) \\
&= \sum_{l=0}^{B-1}(U^l(R)F^l_r)^TS^l(\theta, \phi)
\end{aligned}
\label{eq:real_fourier_transform_gr_simplified}
\end{equation}
where, $U^l(R)=\overline{T^l}D^l(R)(T^l)^T$. $\overline{T^l}$ is the complex conjugate of $T^l$ and $D^l$ is the Wigner D-matrix. The expansion of \eqref{eq:real_fourier_transform_gr_simplified} is possible as rotations are expressed as Wigner D-Matrices in the spectral domain and applying a rotation to the basis functions $S^l$ is equivalent to applying a linear transformation of the basis functions by the equivalent Wigner D-Matrix. From \eqref{eq:real_fourier_transform_gr} and \eqref{eq:real_fourier_transform_gr_simplified}, it can be noticed that $G^l_r = U^l(R)F^l_r$. Thus, $G^l_r$ is obtained by applying the transformation $U^l(R)$ to the Fourier coefficients of $f_r$. More details on commonly used properties of the Winger D-matrix can be found in \cite{lee2018real, blanco1997evaluation, kostelec2008ffts}. 
The goal of the rotation estimation is to find $R$ given $f_r$ and $g_r$.

\subsubsection{Correlation over $\bm{SO}(3)$ and its derivatives}
The correlation of $f_r$ and $g_r$ on $\bm{SO}(3)$ is computed as:
\begin{equation}
\mathcal{C}_r(R) = corr(f_r,g_r) = \frac{1}{4\pi}\sum_{l=0}^{B-1}(G^l_r)^TU^l(R)F^l_r
\label{eq:correlation_rotation}
\end{equation}

This result is obtained after simplification, by replacing $f_r$ and $g_r$ by their Fourier representations expressed in~\eqref{eq:real_fourier_transform_fr} and~\eqref{eq:real_fourier_transform_gr_simplified}, using the convolution theorem of the Fourier transform\footnote{convolution in the spatial domain is equivalent to the multiplication of the Fourier coefficients in the spectral domain}, and the orthogonality principle of basis $S^l$. The relation $\langle S^l(\theta, \phi), (S^l(R^T(\theta, \phi))^T)\rangle=\frac{1}{4\pi}U^l(R)$ directly results from the orthogonality of the basis vectors $S^l$, where the operation $\langle . \rangle$ is the inner product on $\mathcal{L}^2(SO(3))$. In~\eqref{eq:correlation_rotation}, only $U^l$ depends on the rotation $R$, so the derivative of $\mathcal{C}_r$ can be obtained by computing the derivative of $U^l$. The derivative of $U^l$ at $R$ with respect to an elementary rotation $R_{\epsilon} = \exp(\epsilon \hat{\eta})$ ($\epsilon \approx 0$ and $\eta \in \mathbb{R}^3$) is computed as:
\begin{equation}
\left.\frac{d}{d \epsilon}\right|_{\epsilon=0} U^l(R \exp (\epsilon \hat{\eta}))= U^l(R)\left.\frac{d}{d \epsilon}\right|_{\epsilon=0} U^l(\exp (\epsilon \hat{\eta}))
\label{eq:ul_derivative}
\end{equation}

In the previous equation, the homomorphism property of $U^l$ is used, \emph{i.e}, $U^l(R_1R_2) = U^l(R_1)U^l(R_2)$ for $R_1, R_2 \in \bm{SO}(3)$. Then, the derivative of $\mathcal{C}_r$ is then computed by:
\begin{equation}
\begin{aligned}
\left.\frac{d}{d \epsilon}\right|_{\epsilon=0} \mathcal{C}_r(\exp(\epsilon\hat{\eta})) &= \frac{1}{4\pi}\sum_{l=0}^{B-1}(G^l_r)^TU^l(R)u^l(\eta)F^l_r \cdot \eta \\
&=\nabla \mathcal{C}_r(R, \eta) \cdot \eta
\end{aligned}
\label{eq:pre_cr_derivative}
\end{equation}
where, $\nabla \mathcal{C}_r(R, \eta) \in \mathbb{R}^3$ is the gradient of $\mathcal{C}_r(R)$ around the axis $\eta$ and $u^l(\eta) = \left.\frac{d}{d \epsilon}\right|_{\epsilon=0} U^l(\exp (\epsilon \hat{\eta}))$. Evaluating the gradient $\nabla \mathcal{C}_r(R, \eta)$ at $\eta = e_x, e_y, e_z$ allows finding the elementary rotation which applied to $R$ increases the correlation $\mathcal{C}_r$. More formally:
\begin{equation}
\left.\nabla \mathcal{C}_r(R, e_k)\right|_{k\in \{x, y, z\}} = \frac{1}{4\pi}\sum_{l=0}^{B-1}(G^l_r)^TU^l(R)u^l(e_k)F^l_r
\label{eq:cr_derivative}
\end{equation}

The computation of $u^l(e_k)$ is straightforward as it is a direct differentiation of the entries of the Wigner D-matrix for which an analytic derivative can be computed as in~\cite{lee2018real}. 

We can now use a gradient ascent method to iteratively find the ideal rotation. The following cost-function can be formulated to evaluate the performance of the rotation estimation algorithm on $\bm{SO}(3)$:
\begin{equation}
J_r(R) = \frac{1}{2}||g_r(\theta, \phi) - f_r(R^T(\theta, \phi))||^2
\label{eq:rotation_cost_function}
\end{equation}
%
%
%
\subsection{Controller}
To estimate the transformation $H = (R, T) \in \bm{SO}(3)\times\mathbb{R}^3$ between current and reference point clouds, the control law given in ~\eqref{eq:control_law} is used.
\begin{equation}
\begin{aligned}
T &= T + \lambda_t\nabla_{glob} T \\
R &= R\exp{(\lambda_r \widehat{\nabla \mathcal{C}_r})}
\end{aligned}
\label{eq:control_law}
\end{equation}
where, $\lambda_t, \lambda_r \in \mathbb{R}^+$ and $\lambda_t, \lambda_r <1$. $\nabla_{glob} T$ and $\nabla \mathcal{C}_r$ are computed from~\eqref{eq:fourier_shift_solution} and \eqref{eq:cr_derivative}, respectively. At the first iteration, $R$ and $T$ can be initialised randomly or set to identity and zero. 
The controller converges when
\begin{equation}
    ||\nabla_{glob} T|| + ||\nabla \mathcal{C}_r|| < \epsilon_g
    \label{eq:updaterule}
\end{equation}
with $\epsilon_g \in \mathbb{R}^+$ being the tolerance. 
For commanding the robot, the following control law is used 
\begin{equation}
    \dot {\mathbf{q}} = \mathcal{\bm{J}}_c^+\dot{X_c}
    \label{eq:robot_controller}
\end{equation}
with $\mathcal{\bm{J}}_c^+$ being the robot Jacobian pseudoinverse expressed in the camera frame, $\dot{{\mathbf{q}}}$ the vector of robot joint velocities and $\dot{X_c}$ the camera velocities, derived from \eqref{eq:control_law}. The complete control algorithm is presented in Alg.~\ref{alg:algorithm}.
%
%
\begin{algorithm}[t]
\label{alg:algorithm}
 \caption{3D spectral domain visual servoing}
  Initialise $R$ to identity\\
  Initialise $T$ to zero\\
  Initialise the step sizes $\lambda_t, \lambda_r$ and the tolerance
  $\epsilon_g$ \\
  Compute $f_t$ \eqref{eq:voxel}, $f_r$ \eqref{eq:egi}, $F_t$ \eqref{eq:fourier_coefs}, $F^l_r$ of the target cloud \\
  \While{$||\nabla_{glob} T|| + ||\nabla \mathcal{C}_r|| >= \epsilon_g$}{
  Capture the scene point cloud
  Compute $g_t$ \eqref{eq:voxel}, $g_r$ \eqref{eq:egi}, $G_t$ \eqref{eq:fourier_coefs}, $G^l_r$ of the reference cloud \\
  Compute $\nabla_{glob} T$ \eqref{eq:fourier_shift_solution} and $\nabla \mathcal{C}_r$ \eqref{eq:cr_derivative} \\
  Apply the update rule \eqref{eq:control_law} \\
  Compute the cost $J = J_t(T) + J_r(R)$ \\
   Command the robot using \eqref{eq:robot_controller}\\
  }
  Get final transformation $H = (R, T)$ \\

\end{algorithm}
%

%

%
\section{Experimental Validations}
\label{sec:results}
\subsection{Setup Description}
Experiments are performed using camera-acquired point clouds. Two different tests are shown in this work. We first validate the model alignment process, where a full reference model of an object is converged onto a scene point cloud. Next, we show direct visual servoing tests conducted using a 7-axis cobot (KUKA iiwa) fitted with a wrist-mounted depth camera (Ensenso N35). For this case, the entire point cloud is used to position the robot at a target location. The point cloud processing and controller software are developed in C++ and are executed from a PC running Windows with Intel i7 4 core CPU with $2.9~\mathrm{GHz}$ frequency. Point cloud library (PCL) \cite{Rusu_ICRA2011_PCL} is used for point cloud processing, and FFTSO3 \cite{lee2018real} and FFTW \cite{frigo1998fftw} libraries are used for spectral analysis.  

As mentioned earlier in Sec. \ref{sec:intro}, the reference object point cloud for model alignment experiments are built offline by stitching multiple clouds as in \cite{marturi2019dynamic}. Note that the point normals are obtained directly at the time of cloud acquisition. The point clouds are voxelized with a grid of resolution $8~\mathrm{mm}$, while the surface normals are discretised on the unit sphere with a bandwidth of $B=16$. The maximum degree of expansion of the harmonic coefficients on the sphere is $l_{max} = 32$. These values are estimated empirically and provide good performance in terms of speed and alignment accuracy. The three main factors that control the convergence speed of our approach are the voxel grid resolution, the EGI bandwith, and the parameters $\lambda_t$ and $\lambda_r$. Finer grids require computing higher number of Fourier coefficients, which in turn slows down the process. With the aforementioned parameters, on an average, the current processing speed of our approach is $8.7~\mathrm{ms/iteration}$.
\begin{figure}
    \centering
    \includegraphics[width=\columnwidth]{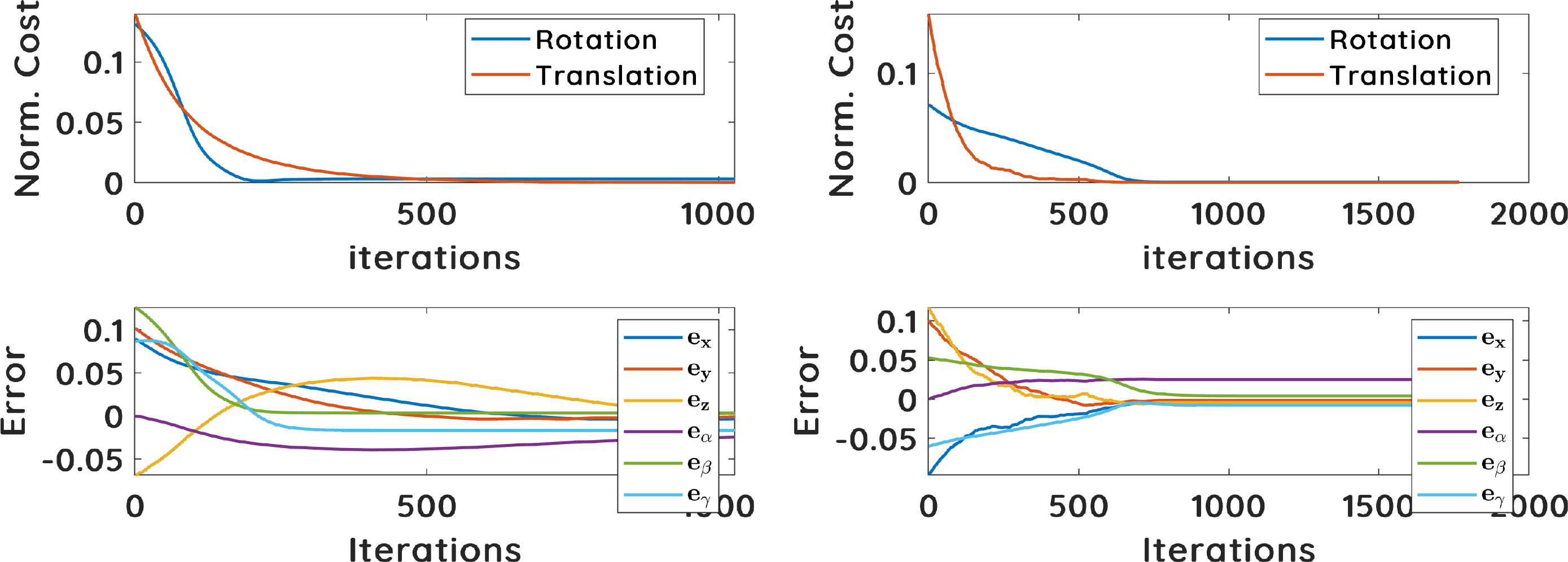}
    \caption{Convergence plots for the (left) mug and (right) gas knob objects shown in Fig. \ref{fig:fig1}.}
    \label{fig:fig1_plots}
\end{figure}
\subsection{Model Alignment Analysis}
\begin{figure}
    \centering
    \includegraphics[width=\columnwidth]{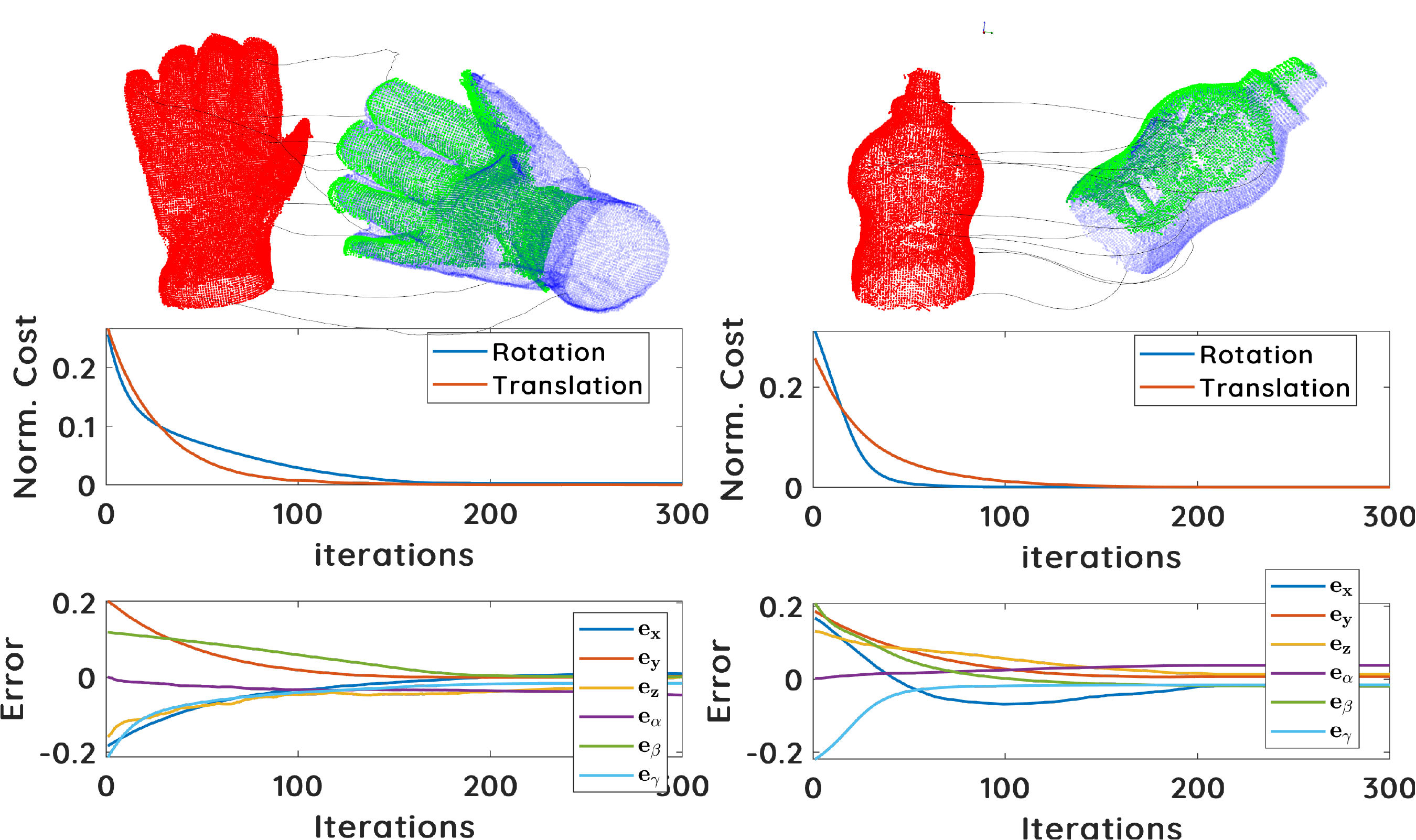}
    \caption{Model alignment analysis in case of a full reference model being aligned to a partially observed scene. Red, blue and green clouds respectively represent, reference, aligned and target. Results for two objects are shown (left) glove and (right) mustard can.}
    \label{fig:align_res}
\end{figure}

\begin{figure}
    \centering
    \includegraphics[width=\columnwidth]{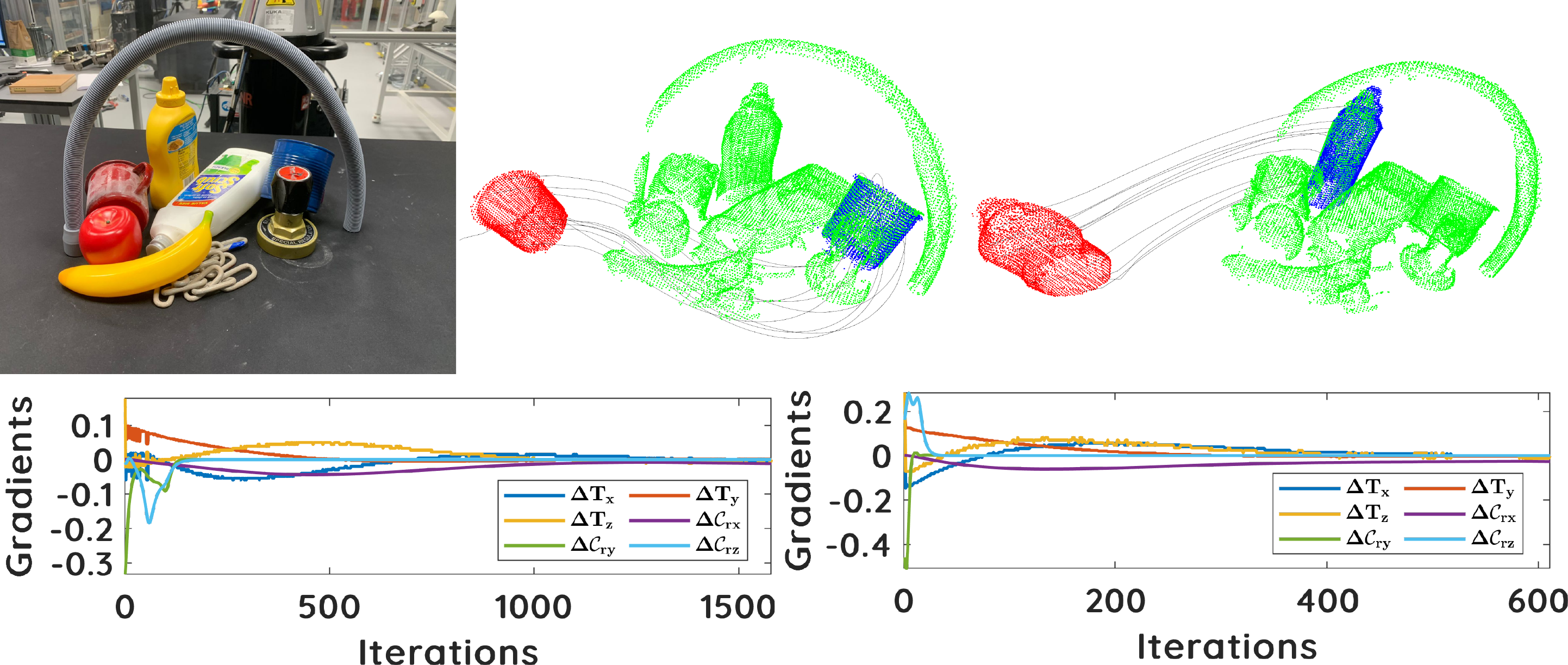}
    \caption{Model alignment analysis in case of cluttered scenes. (top) The used clutter scene and alignment for two different objects are shown; and  (bottom) corresponding plots showing the evolution of translation and rotation gradients during the alignment.}
    \label{fig:clutter_res}
\end{figure}

The following three different experiments are conducted validating the model alignment ability of our approach: (C--1) a full single object model cloud is aligned on to its arbitrarily transformed version (Fig. \ref{fig:fig1} and Fig. \ref{fig:fig1_plots}); (C--2) a full model cloud is aligned on to a partially observed target cloud (Fig. \ref{fig:align_res}); and (C--3) a full model cloud is aligned onto a cluttered scene of objects (Fig. \ref{fig:clutter_res}). Note that for all these experiments, both the reference and target clouds share the same global frame. 
Different household objects are used for the tests and the clutter scenes are built by randomly positioning these objects as shown in Fig. \ref{fig:clutter_res}. The result images shown in Fig. \ref{fig:fig1}, \ref{fig:fig1_plots}, \ref{fig:align_res}, and \ref{fig:clutter_res},  show sample screenshots during the alignment process and the evolution of costs and errors. In case of clutter, evolution of gradients is shown as the target location, \textit{i.e.}, the true ground truth position, of the object being matched is not known beforehand. Detailed results can be seen in the supplementary video.

\midsepremove
\begin{table}
    \centering
    \caption{Average final values during model alignment.}
    \label{tab:cost_table}
    \begin{threeparttable}
    \begin{tabularx}{\columnwidth}{>{\raggedright\arraybackslash}X|
 >{\hsize=1\hsize\centering\arraybackslash}X|
 >{\hsize=1\hsize\centering\arraybackslash}X|
 >{\hsize=1\hsize\centering\arraybackslash}X}
         \toprule
         &  C--1 (cost) & C--2 (cost) & C--3 (gradient) \\
         \midrule
        Trans. error\tnote{1} & 1.77524e-5 & 8.11356e-5 & 6.4e-05  \\
        Rotation error\tnote{1} & 2.747765e-2 & 3.2e-2& 1.199277e-12  \\
        \bottomrule
    \end{tabularx}
    \begin{tablenotes}
        \item[1] \footnotesize{Computed as average of $(\mathrm{real} - \mathrm{estimated})^2$. Note that this applies only to C--1 and C--2.}
    \end{tablenotes}
    \end{threeparttable}
\end{table}

From the obtained results, it can be seen that the model clouds are successfully aligned with the target clouds for all the test cases. The average final convergence costs for conditions C--1 and C--2, and the average final gradient change for C--3 are shown in the Table \ref{tab:cost_table}. This clearly demonstrates the accuracy of our approach in terms of convergence and model alignment. Furthermore, our method showcased superior performance in case of the complex conditions like matching a model to partially observed data as well as to extremely unstructured (and occluded) scenes containing a heap of objects. A good convergence and model alignment are observed even for these complex cases.
\subsection{Robot Positioning Tests: Direct Visual Servoing}
\begin{figure}
    \centering
    \includegraphics[width=\columnwidth]{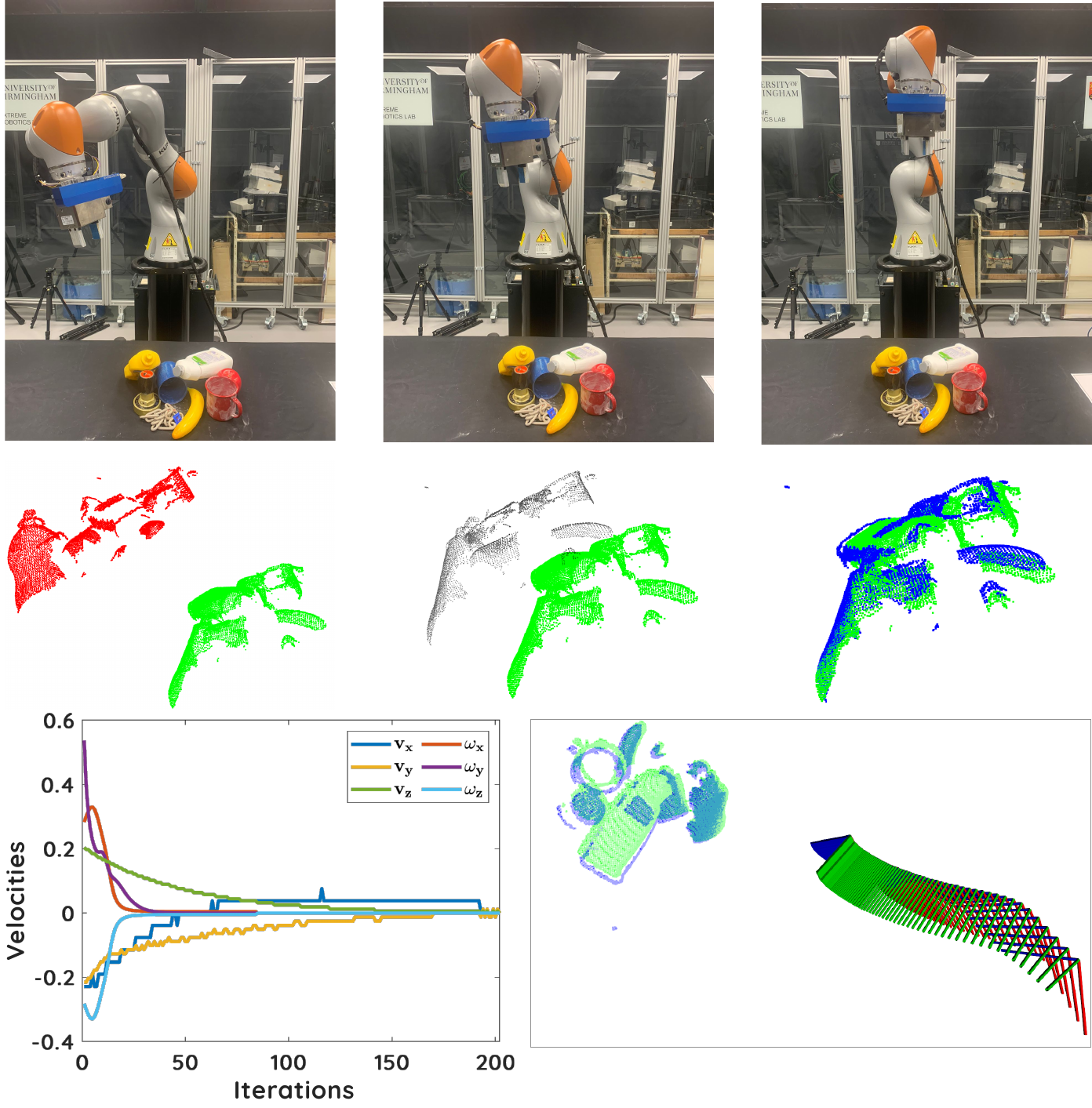}
    \caption{Illustration of direct visual servoing. Top row shows the robot at starting, intermediate, and final positions. Middle row shows the initial, intermediate and final point clouds. Bottom row shows the task convergence plot and the trajectory followed by the robot end-effector. For this experiment, the entire point cloud is used instead of any local object model matching. See supplementary video for more details.}
    \label{fig:dvs_result}
\end{figure}
For this test, we consider a planar positioning task where the robot end-effector’s position is controlled in a direct visual servoing fashion. As stated earlier, the entire point cloud is used to generate the robot control commands, \textit{i.e.}, without using any local model matching. This test has been performed with a cluttered scene to test the ability of our approach in case of challenging conditions. For the sake of demonstration, the reference cloud is acquired at the robot home position. After this, the robot is moved to a random position in the task space from where the visual servoing starts. Note that all joints of the robots are moved to ensure large transformation between the current and reference positions. Results obtained with this experiment are depicted in the Fig. \ref{fig:dvs_result}. From these results, it is clearly evident that the method performs well while matching an entire cloud starting from a position where only a part of it is visible. The convergence plots demonstrate the smooth motion of the robot in reaching a target position. Detailed results can be found in the supplementary video.
%
\section{Conclusion}
\label{sec:conclusion}
In this paper, we presented a visual servoing method based on the 3D spectral domain registration using point clouds. The presented approach makes use of the Fourier shift property to estimate translations, and the real spherical harmonics on $\bm{SO(3)}$ to estimate rotations, in order to iteratively align a reference point cloud on to a target one. This methodology has been initially used to align 3D models on to different scenes represented as point clouds, and later used to control the end-effector position of a robot arm to accomplish a planar positioning task. The obtained experimental results demonstrate the efficiency of our approach in terms of aligning a 3D model under complex conditions as well in positioning  the robot arm. 

Future work is concentrated on using the proposed approach and perform a full scale 3D object manipulation involving static and moving object grasping \cite{adjigble2021spectgrasp,de2021dual}.
\bibliographystyle{IEEEtran}
\bibliography{IEEEabrv,references}
\end{document}